%% file: arxiv_version.tex
\documentclass[10pt,twocolumn,letterpaper]{article}
\input{math_commands.tex}

\usepackage{cvpr}              

\usepackage{graphicx,adjustbox}
\usepackage{url}
\usepackage[linesnumbered,ruled,vlined]{algorithm2e}
\SetKwComment{Comment}{/* }{ */}
\SetAlFnt{\small}
\SetKwInput{kwInput}{Input}
\SetKwInput{kwOutput}{Output}
\SetKwBlock{DataGeneration}{DataGeneration($\mathcal{S}^{t-1}, \mathcal{P}$)}{}
\SetKwBlock{ClientUpdate}{ClientUpdate($\mathcal{C}^t_k, \mathcal{S}^{t-1}$, $\mathcal{M}^{t-1}, \mathcal{P}$)}{}
\RestyleAlgo{ruled}
\usepackage{amsmath}
\usepackage{array}
\usepackage{multirow}
\usepackage{booktabs}
\usepackage[table]{xcolor}
\usepackage[normalem]{ulem}
\useunder{\uline}{\ul}{}
\usepackage{float, wrapfig}
\usepackage{resizegather}
\usepackage{enumitem}
\usepackage[font=small,skip=0pt]{caption}
\def\code#1{\texttt{#1}}
\usepackage{bbm}
\usepackage[accsupp]{axessibility}
\usepackage{amssymb}

\usepackage[pagebackref,breaklinks,colorlinks]{hyperref}
\usepackage[capitalize]{cleveref}
\crefname{section}{Sec.}{Secs.}
\Crefname{section}{Section}{Sections}
\Crefname{table}{Table}{Tables}
\crefname{table}{Tab.}{Tabs.}

\def\cvprPaperID{14025} 
\def\confName{CVPR}
\def\confYear{2024}
\begin{document}

\title{Text-Enhanced Data-free Approach for Federated Class-Incremental Learning}

\author{Minh-Tuan Tran$^1$, Trung Le$^1$, Xuan-May Le$^2$, Mehrtash Harandi$^1$, Dinh Phung$^{1,3}$\\\
$^1$Monash University,  $^2$University of Melbourne, $^3$VinAI Research\\
{\tt\small \{tuan.tran7,trunglm,mehrtash.harandi,dinh.phung\}@monash.edu} \\ \tt\small xuanmay.le@student.unimelb.edu.au}
\maketitle

\begin{abstract}

Federated Class-Incremental Learning (FCIL) is an underexplored yet pivotal issue, involving the dynamic addition of new classes in the context of federated learning. In this field, Data-Free Knowledge Transfer (DFKT) plays a crucial role in addressing catastrophic forgetting and data privacy problems. However, prior approaches lack the crucial synergy between DFKT and the model training phases, causing DFKT to encounter difficulties in generating high-quality data from a non-anchored latent space of the old task model. In this paper, we introduce LANDER (Label Text Centered Data-Free Knowledge Transfer) to address this issue by utilizing label text embeddings (LTE) produced by pretrained language models. Specifically, during the model training phase, our approach treats LTE as anchor points and constrains the feature embeddings of corresponding training samples around them, enriching the surrounding area with more meaningful information. In the DFKT phase, by using these LTE anchors, LANDER can synthesize more meaningful samples, thereby effectively addressing the forgetting problem. Additionally, instead of tightly constraining embeddings toward the anchor, the Bounding Loss is introduced to encourage sample embeddings to remain flexible within a defined radius. This approach preserves the natural differences in sample embeddings and mitigates the embedding overlap caused by heterogeneous federated settings. Extensive experiments conducted on CIFAR100, Tiny-ImageNet, and ImageNet demonstrate that LANDER significantly outperforms previous methods and achieves state-of-the-art performance in FCIL. The code is available at \url{https://github.com/tmtuan1307/lander}.

\end{abstract}
\section{Introduction}
\label{sec:intro}
Federated Learning (FL) is a decentralized and privacy-preserving technique enabling collaboration among diverse entities, such as organizations or devices \cite{flcm, flniid, fls1, fls2}. In FL, multiple users (clients) train a common (server) model in coordination with a server without sharing personal data. FL in has recently gained attention in various fields like healthcare \cite{fl_hc}, IoT \cite{fl_iot}, and autonomous driving \cite{fl_ad}. While conventional FL studies assume static data classes and domains, the reality is that new classes can emerge, and data domains can change over time \cite{cl1, cl2, cl3, cl4}. For example, \cite{fcl_rainbow} reveals shifting customer interests in an online store with seasons, and \cite{fclwt} discusses the need for healthcare models to adapt to detect new diseases. Handling continuously emerging data classes through entirely new models is impractical due to substantial computational resources. Alternatively, transfer learning from pre-existing models may be considered, but it faces the issue of catastrophic forgetting \cite{forget1, forget2}, degrading performance on previous classes.

To tackle catastrophic forgetting in FL, recent studies \cite{fcl, fclvi, fclkd, fclwt} propose the concept of Federated Continual Learning (FCL) which incorporate Continual Learning (CL) principles \cite{cl1, cl2, cl5, cl6} into Federated Learning. In that, the most popular setting is Federated Class-Incremental Learning (FCIL) \cite{mfcl,target,fedcil,fclwt,fcl}, providing the flexibility to add new classes at any time. 

The key challenges in FCIL are to mitigate catastrophic forgetting and ensure data privacy. To tackle these challenges, the common FCIL process unfolds in two main phases: client/server (model) training and the Data-Free Knowledge Transfer (DFKT) phase. In this context, unlike classic knowledge distillation methods \cite{kd4,kd5,se,uda}, DFKT \cite{dfkd_dnn, adi, cmi, nayer} has emerged as a pivotal technique since it can transfer the knowledge from the last task model (as a teacher) to the current model (as a student) to mitigate the forgetting problem but without accessing raw training data, thus ensuring privacy. The core idea behind DFKT is to generate synthetic data with confident predictions from the teacher. In other words, these synthetic data points reside in the high-confidence region of the teacher's predictions. Subsequently, this synthetic data is utilized to train the student model, effectively addressing the issue of forgetting.

However, the utilization of DFKT in previous methods has frequently resulted in unsatisfactory outcomes \cite{target,mfcl,fedcil}. This can be attributed to the fact that existing methods lack anchors shared between the client and teacher/server models to constrain high-confidence regions of client and server models, making them more well-organized and thereby facilitating the generation of synthetic data. Consequently, due to the disorganized and non-anchored high-confidence regions of the teacher models, to cover all knowledge from the teacher, previous methods need to generate a large amount of synthetic data in these complex regions, including both high/low-quality images, limiting the effectiveness of DFKT in mitigating catastrophic forgetting. For instance, numerous existing cat images could confidently be classified as belonging to the dog class by any teacher model, as discussed in \cite{at1, at2, at3}.

In this paper, we address the problem by introducing a novel method named LANDER (\underline{LA}bel text ce\underline{N}ter \underline{D}ata-free knowledg\underline{E} transfe\underline{R}). LANDER leverages the label-text embedding (LTE) produced by pretrained language models to reorganize the latent embedding of training data, facilitating the synthesis of high-quality samples in DFKT. Specifically, our method treats the LTEs as the anchors and optimizes the feature embeddings of training samples around these LTEs area, ensuring this area contains more semantically meaningful information. Importantly, our method queries the LTEs from the language model only once. This LTEs is stored in memory for subsequent processing, and we do not involve the language model in the training process. During the DFKT phase, LANDER capitalizes on these advantages by generating samples in proximity to these LTEs, thereby creating the synthetic data with more valuable features. This approach mitigates catastrophic forgetting, as these more meaningful samples help the model retain knowledge from previous tasks. LANDER departs from conventional approaches by using LTE as input to the generator, shifting the source of randomness from the input level to the layer level, resulting in faster and more diverse sampling. Furthermore, we introduce the concept of the Bounding Loss to encourage sample embeddings to remain flexible within a defined radius, rather than attempting to make them as close as possible to the LTEs. This method retains inherent differences in embeddings and alleviates overlap arising from heterogeneous federated settings. Our contributions can be summarized as follows:
\begin{itemize}[noitemsep, nolistsep]
    \item We propose LANDER which leverage the power of pretrained language models in FCIL. It utilizes label text embeddings as anchors to enhance knowledge transfer from previous models to the current model.
    \item We propose preserving natural embedding differences with the Bounding Loss to address overlap issues in imbalanced federated settings.
    \item We enhance data privacy by introducing a learnable data stats for the data-free generator, eliminating the need for clients to disclose specific data information.
    \item Extensive experiments on CIFAR100, Tiny-ImageNet, and ImageNet show that LANDER outperforms previous methods, establishing itself as the state-of-the-art (SOTA) solution for FCIL.
\end{itemize}

\section{Related Work}

\noindent
\textbf{Data-free Knowledge Transfer.} DFKT or data-free knowledge distillation is an approach that enables knowledge transfer from a teacher model to a student model without the need for training data. Recent techniques, such as those introduced in \cite{zadf, zskt}, utilize a generative model to synthesize images, guiding the teacher's predictions. The student and generator undergo joint adversarial training, facilitating rapid exploration of synthetic distributions. Notably, the use of a pretrained text encoder in DFKT, as demonstrated in \cite{nayer}, has achieved SOTA results. DFKT is gaining popularity in CL \cite{dfkd_dnn, rdfcil} and FL \cite{dense, dfrd} due to its ability to preserve knowledge without relying on memory while addressing privacy concerns.

\noindent
\textbf{Continual Learning.} Catastrophic forgetting, a significant challenge in machine learning \cite{forget2}, occurs when a model's performance on previously learned data decreases as it is trained on new examples. CL \cite{cl_first} addresses this issue by enabling models to acquire new knowledge while retaining existing knowledge. Strategies include regularization terms \cite{cl_rt, cl_rt2, cl_rt3}, experience replay \cite{cl_er, cl_er2, cl_er3}, generative models \cite{cl_gr, cl_gr2, cl_gr3}, and isolating architectural parameters \cite{cl_iap, cl_iap2, cl_iap3}. DFKT methods \cite{deepdream,adi,rdfcil} prove promising for CL, especially in privacy-sensitive applications. CL encompasses various learning scenarios: task-incremental learning (TIL), domain-incremental learning (DIL), and class-incremental learning (CIL) \cite{cl_3s}. TIL involves separate tasks, DIL maintains a consistent output space, and CIL gradually introduces new tasks and classes.

\noindent
\textbf{Federated Continual Learning.} In real-world scenarios, local user data constantly evolves due to changing interests or data loss concerns. Federated continual learning (FCL) addresses the challenge of updating global models with evolving user data while retaining previous knowledge. Key contributions include \cite{fclwt}, focusing on TIL with distinct task IDs, and \cite{fclkd}, which employs knowledge distillation. Additionally, \cite{fcl} diverges by assuming clients have sufficient memory for storing and sharing data. Other studies, like \cite{fedspeech, fedcm, fedps}, explore FCL in diverse domains.

\begin{figure*}[t]
\begin{center}
\includegraphics[width=0.85\linewidth]{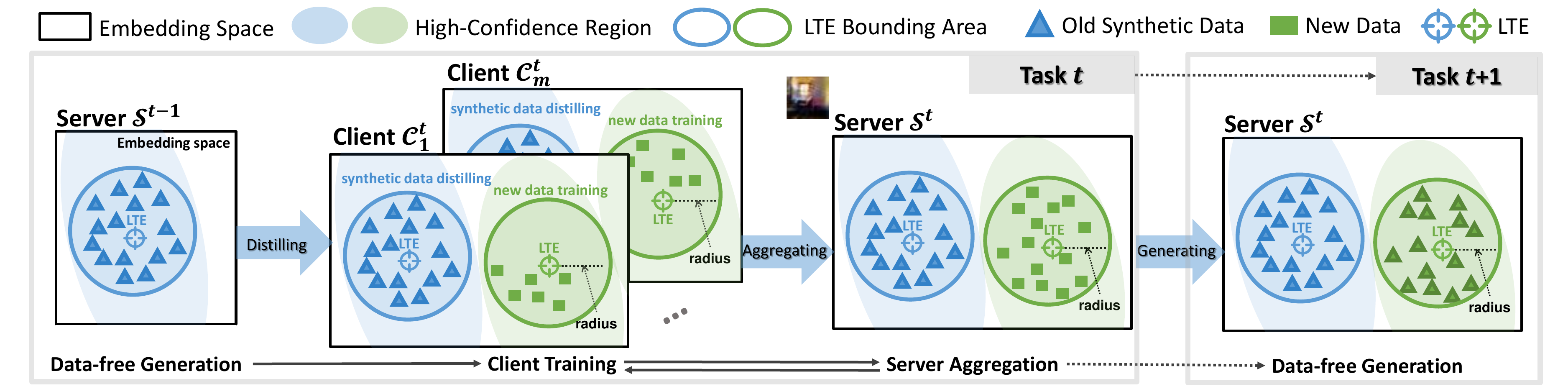}
\end{center}
\caption{LANDER's motivation centers on using shared LTE as a key constraint for new task features and applying feature distillation from the previous server to anchor and organize the latent space of the client/server model. Subsequently, generating samples around the shared anchor LTE facilitates efficient data-free knowledge transfer from the previous to the current task.}
\label{fig:motivation}
\end{figure*}

\section{Federated Class-Incremental Learning}

In our paper, our primary focus is on Federated Class-Incremental Learning, which applies class-incremental learning in a federated setting. FCIL framework comprises a global server $\mathcal{S}$ and multiple clients ($\mathcal{C}_1,\cdots,\mathcal{C}_m$). In that, each client is trained to address a sequence of distinct and non-overlapping tasks without sharing their data with one another or with the central server to protect their data privacy. For task $t$, we denote $\mathcal{D}^t_k$ ($1\leq k \leq m$) as the data set of task $t$ of the client $k$, consisting of $N^t_k$ pairs of samples and their labels $\{ (\vx_i^{kt}, \vy_i^{kt})\}_{i=1}^{N^t_k}$. Labels $\vy_i^t$ belong to non-overlapping subsets of classes $\mathcal{Y}^t_k \subset \mathcal{Y}$, where $\mathcal{Y}$ is the set of all possible classes. To provide clarity, we denote the server and clients model for task $t$ as $\mathcal{S}^t$ and ($\mathcal{C}_1^t,\cdots,\mathcal{C}_m^t$), respectively. The primary objective of FCIL is to maintain a global model performance in previous and current tasks. In our privacy-conscious scenario, the task sequence is presented in an undisclosed order, and each client $\mathcal{C}^t_k$ can exclusively access its local data $\mathcal{D}^t_k$ for task $t$ during that task's training period, with no further access allowed thereafter. Note that the models are trained in a distributed setting, with each party having access to only a subset of the classes $\mathcal{Y}^t$ (\ie, non-IID). 

FCIL has recently gained significant attention for its challenging nature and its closer alignment with real-world scenarios, especially within the context of federated learning. It's worth noting that in most FL applications, task IDs are not readily available, and the preferred approach is to train a unified model that accommodates all observed data. Several methods have been proposed recently to address this problem. For example, FedCIL \cite{fedcil} suggests local training of the discriminator and generator to leverage generative replay, effectively compensating for the absence of old data and mitigating forgetting. On the other hand, MFCL \cite{mfcl} and TARGET \cite{target} introduce a data-free approach in which the generative model is trained by the server. This approach reduces client training time and computational requirements while still eliminating the need for access to their private data. 

It is clear that DFKT plays a pivotal role in mitigating the issue of catastrophic forgetting and ensuring privacy preservation in the majority of current FCIL methods. However, existing methods including TARGET \cite{target}, MFCL \cite{mfcl} and FedCIL \cite{fedcil} treat DFKT as standalone modules, without integrating it into the training phases. This isolated approach hampers their effectiveness in mitigating the model forgetting problem.

\section{Our Proposed Method: LANDER}
\subsection{Motivations of LANDER}

For data-free FCIL, we follow the framework at \cite{target, mfcl}. For task $t$, we firstly perform DFKT over the server model up to task $t-1$ (i.e., $\mathcal{S}^{t-1}$) to generate synthetic data $\mathcal{M}^{t-1}$. Afterward, the server must engage in communication with the clients for $c$ rounds, where each round comprises two phases: client-side training and server aggregation. In client-side training, given the client $k$, its model $\mathcal{C}^t_{k}$ is trained using the new task data $\mathcal{D}^t_{k}$ and the old task synthetic data $\mathcal{M}^{t-1}$. In server aggregation, the client models are then sent to the server side to aggregate the server model (i.e., $\mathcal{S}^t = \frac{1}{m}\sum_{k=1}^m\mathcal{C}^t_k$).

However, this naive mechanism lacks synergy between the clients and the server to constrain the complexity of high-confidence regions of the server model $\mathcal{S}^{t-1}$ in order to facilitate DFKT and generate more qualified synthetic old task images.


To organize the high-confidence regions of the client models and server model more effectively, we propose imposing constraints during the training of the client models. Specifically, given the client $k$, (i) for new data in $\mathcal{D}^t_{k}$, the feature vectors of $(\vx, \vy) \in \mathcal{D}^t_k$ at the penultimate layer of the client model $\mathcal{C}^t_k$ must center around the LTE of the class $\vy$, and (ii) the feature vectors of old-task synthetic data $(\hat{\vx}, \hat{\vy}) \in \mathcal{M}^{t-1}$  of the client model $\mathcal{C}^t_k$ must distill those of the server model $\mathcal{S}^{t-1}$. This approach aims to organize the high-confident regions for classes of the client models more coherently around meaningful LTEs.

Moreover, on the server side, we aggregate the client models to obtain the server model. Hence, the well-organized high-confident regions for classes of the client models are inherited by the server model. Using LTE as the anchor, this inherited property certainly facilitates the server model to be data-freely distilled more effectively for generating more qualified images for old tasks, thereby alleviating catastrophic forgetting.

Figure \ref{fig:motivation} illustrates the motivations of our proposed LANDER. In particular, the well-organized high-confident regions for classes of the client models are inherited by the server model through the aggregation operation, further facilitating the DFKT phase for generating high-quality images of the old tasks.

\begin{figure}[t]
\begin{center}
\includegraphics[width=1\linewidth]{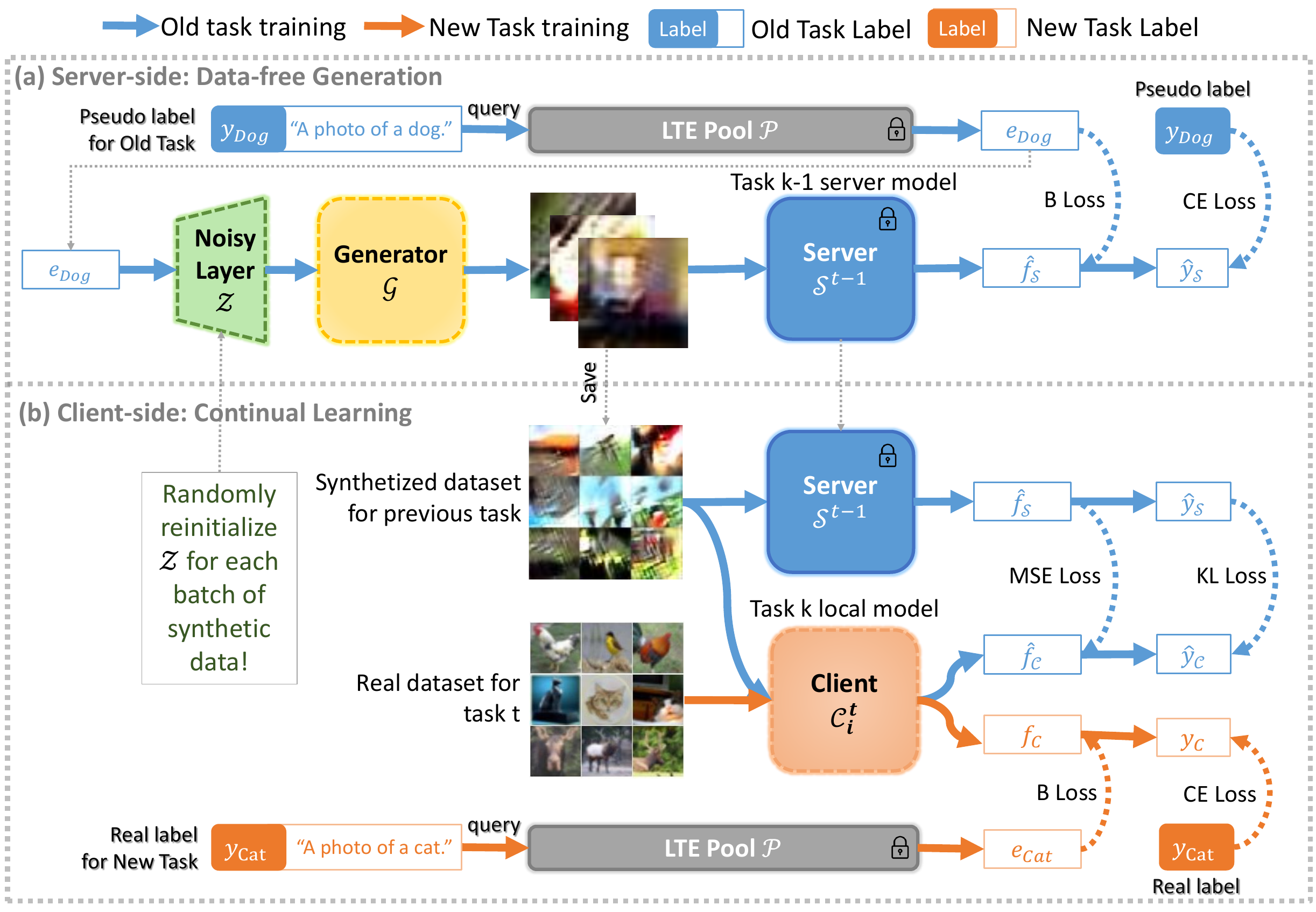}
\end{center}
\caption{General Architecture of LANDER: (a) We utilize the previous server model (trained on task $t-1$) to synthesize the data. (b) Subsequently, we use this data to train the $k$-th task. We use the LTE as the anchor to constrain the features in both the client and generator, enhancing performance.}
\label{fig:lander}
\end{figure}

In the next section, we first discuss how to effectively organize the high-confidence regions of the client/server model (Section \ref{sec:ltc}). Then, we delve into the details of the two main components of our method: Client-Side Training (Section \ref{sec:c1}) and Server-Side Data Generation (Section \ref{sec:s1}). The overall architecture of our method is illustrated in Figure \ref{fig:lander}, while Algorithm \ref{alg:lander} provides a comprehensive overview of the entire training process for LANDER.

\begin{algorithm}
\caption{LANDER}
\label{alg:lander}
\kwInput{$E$: local epoches, $I$: generation rounds, $g$ generator training steps, $T$: number of tasks, $c$: number of communication rounds, synthetic data $\mathcal{M}^0 =  \{\} $, LTE pool $\mathcal{P}$ = \{\}.}
\ForEach{each task $t = 1,\cdots,T$}{
    Store all embeddings $\ve_{{\vy}} = \mathcal{E}(Y^t_{{\vy}})$ into $\mathcal{P}$\;
    \lIf{$t \neq 1$}{
        $\mathcal{M}^{t-1} = \textbf{DataGeneration}(S^{t-1}, \mathcal{P})$
    }
    \For{$c$ rounds}{
        \ForEach{\text{each client} $i = 1,\cdots,m$}{
            $\mathcal{C}^t_k = \mathcal{S}^t$\;
            $\mathcal{C}^t_k =$ \textbf{ClientUpdate}($\mathcal{C}^t_k, \mathcal{S}^{t-1}$, $\mathcal{M}^{t-1}, \mathcal{P}$)\;
        }
        ${\mathcal{S}^t} = \sum^{m}_{k=1}{\frac{n_k}{n}{\mathcal{C}_k^t}}$\;
    }
}
\ClientUpdate{
    \For{$E$ \textit{epoches}}{
        \ForEach{batch $(({\vx},{\vy}), \hat{\vx}) \sim \mathcal{D}_k^t \cup \mathcal{M}^{t-1}$}{
            Query $\ve_\vy \sim \mathcal{P}$\;
             \If{$t = 1$}{
                Update $\mathcal{C}_k^t$ by minimizing Eq. \ref{eq:cur}
             }\lElse{
                Update $\mathcal{C}_k^t$ by minimizing Eq. \ref{eq:lcl}
             }       
        }
    }
    \textbf{return} ${\mathcal{C}^t_k}$
}
\DataGeneration{
Initialize $\mathcal{G}, {\mathcal{Q}}, \mu, \sigma, \mathcal{M}^{t-1} =  \{\}$\;
\For{$I$ \textit{rounds}}{
    Initializes noisy layer ${\mathcal{Z}}$ and pseudo label $\hat{\vy}$\;
     Query $\ve_{\hat{\vy}} \sim \mathcal{P}$\; 
    \For{$g$ \textit{steps}}{
        $\hat{\vx} = (\mathcal{G}(\mathcal{Z}(\ve_{\hat{\vy}})) - \mu)/\sigma$\;
        Update ${\mathcal{G}}, {\mathcal{Z}}, \mu, \sigma$ by minimizing Eq. \ref{eq:lzg}\;
    }
    $\mathcal{M}^{t-1}= \mathcal{M}^{t-1} \cup \hat{\vx}$\;
    \For{batch $\hat{\vx} \sim \mathcal{M}^{t-1}$}{
        Update $\mathcal{Q}$ by minimizing Eq. \ref{eq:ld}\;
    }
}
\textbf{return} $\mathcal{M}^{t-1}$
}
\end{algorithm}

\subsection{How to Effectively Organize the High-confidence Regions?}
\label{sec:ltc}
We consider a typical client model $\mathcal{C}$ (i.e., $\mathcal{C}_k$ for client index $k$) and a new task data $\mathcal{D}$ (i.e., $\mathcal{D}^t_k$ for some task $t$ and client index $k$). Given a new data/label pair $(\vx, \vy) \in \mathcal{D}$, we extract the feature $\vf_\mathcal{C}$ at the penultimate layer and prediction $\vy_\mathcal{C}$ for $\vx$ of the client model $\mathcal{C}$ (\ie, $\vy_\mathcal{C}, \vf_\mathcal{C} = \mathcal{C}(\vx)$). We now discuss how to establish constraints on $\vf_\mathcal{C}$ to effectively organize the high-confidence regions of $\mathcal{C}$ on $\mathcal{D}$.   


In previous works, the model $\mathcal{C}$ is trained solely by minimizing the cross-entropy loss (CE) between the prediction $\vy_\mathcal{C}$ and the actual label $\vy$ (ie, $\text{CE}(\vy_\mathcal{C},\vy)$). While this approach places the training data $\vx$ in the high-confidence regions of $\mathcal{C}$, it does not impose any constraints on the feature $\vf_\mathcal{C}$ of $\vx$, leading to unsatisfactory organization and scattering throughout the regions. 

This raises the need for common anchors shared between the client and server to both constrain the feature $\vf$ and facilitate the generation of synthetic data. To address this requirement, we propose considering the label-text embedding of $\vy$ as anchors and optimizing the feature embeddings of corresponding training samples around them. In the field of text embedding, a common observation is that text with similar meanings tends to exhibit closer embedding proximity to one another \cite{doc2vec}. Therefore, LTE has the ability to encapsulate useful interclass information, making it an ideal choice for this problem.

Inspired by \cite{nayer}, we initially prompt the label text $Y_{\vy}$ for $\vy$ by adding the text \code{"A photo of a \{class\_name\}"} where \code{\{class\_name\}} represents the label of the class. For example, if the label is \code{"dog"} the prompt will be \code{ "A photo of a dog."} After that, the LTE is calculated using the pretrained LM as follows:
\begin{equation}
    \ve_\vy = \mathcal{E}(Y_{\vy}),\quad \forall \vy \in \mathcal{Y}\;.
\end{equation}
Importantly, the embedding $\ve_{\vy}$ is generated once and then stored in the LTE pool $\mathcal{P}$, remaining fixed throughout the entire training process without any further fine-tuning of the pretrained language model $\mathcal{E}$. So, all that is required here is the text of the label. Then, we can generate the embedding of all labels $\ve_\vy$ globally and send it to the client just once.

To consider the LTE as the anchor, a naive approach is to minimize the Mean Squared Error (MSE) loss between the alignment $\mathcal{W}(\cdot)$ of the feature $\vf_\mathcal{C}$ and the label embedding $\ve_{\vy}$.
\begin{equation}
    \text{MSE}(\vf_\mathcal{C}, \ve_{\vy}) = \big\|\ve_{\vy} - \mathcal{W}(\vf_\mathcal{C}) \big\|^2\;.
\end{equation}

\begin{figure}[t]
\begin{center}
\includegraphics[width=\linewidth]{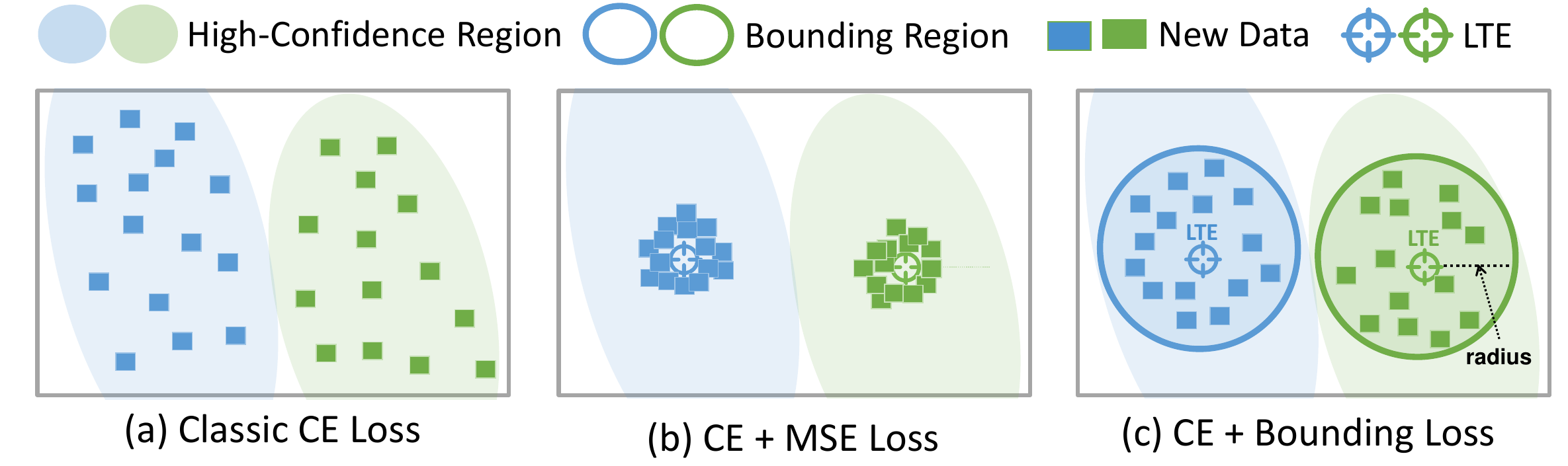}
\end{center}
\caption{Server latent space when using only CE loss;  CE with MSE Loss; and CE with B Loss to constrain the feature embedding. By using our B Loss, the latent features are organized around but still remain flexible within a defined radius $r$ of the LTE center, mitigating the embedding overlap problem.}
\label{fig:bloss}
\end{figure}

\noindent
\textbf{Bounding Loss}. However, in our experiments conducted in Section \ref{sec:ab}, we found that striving to make the feature $\vf_\mathcal{C}$ as close as possible to the LTE $\ve_\vy$ can exacerbate the data imbalance problem in federated learning. This issue arises from the fact that if a client has too few data samples for a particular class, it becomes easy to make the features of these data almost identical to the LTE. This results in embedding overlap between data from different clients, causing the embeddings to be too tight and lacking natural differences for similar classes. To address this problem, we introduce the concept of Bounding Loss (B Loss) (Eq. \ref{eq:b}) to mitigate imbalance issues within heterogeneous federated settings. This approach encourages the embeddings of samples to remain flexible within a defined radius $r$, rather than pushing them to be as close as possible to the anchor LTE. Figure \ref{fig:bloss} illustrates the comparison of our B Loss and other methods. Using B Loss, the LTE is intended to be an anchor point, and each data point should naturally maintain some distance from it due to their inherent differences. This mitigates the overlapping problem and improves performance in heterogeneous federated settings. 
\begin{equation}
    \text{B}(\vf_\mathcal{C}, \ve_{\vy})  = \max\big(0, \big\|\ve_{\vy} - \mathcal{W}(\vf_\mathcal{C}) \big\|^2 - r\big)\;,
    \label{eq:b}
\end{equation}
where a linear projector $\mathcal{W}(\cdot)$ is created to align feature dimensions effectively at a modest cost.

\subsection{Client-Side: LT-Centered Training}
\label{sec:c1}
On the client side of task $t$, the server sends the synthesized data from the previous task, denoted as $\mathcal{M}^{t-1}$, and the previous server model $\mathcal{S}^{t-1}$ to a specific client $k$. Then, we train the local model $\mathcal{C}^t_k$ for task $k$ using both $\mathcal{M}^{t-1}$ and their real training data $\mathcal{D}^t_k$ simultaneously.

For the new task data, we utilize the CE loss to optimize the client model with the real training data and establish constraints on $\vf_\mathcal{C}$ using Bounding Loss at \cref{eq:b} to effectively organize the high-confidence regions of $\mathcal{C}^t_k$ on $\mathcal{D}^t_k$.
\begin{equation}
    \mathcal{L}_{cur} = \text{CE}(\vy_\mathcal{C}, \vy) +  \lambda_{ltc}\text{B}(\vf_\mathcal{C}, \ve_{\vy})\;,
    \label{eq:cur}
\end{equation}
where $\vy_\mathcal{C}, \vf_\mathcal{C}$ are from $\mathcal{C}^t_k(\vx)$, and $(\vx,\vy) \in \mathcal{D}^t_k$.

For the synthetic data, we perform the knowledge distillation on $\mathcal{M}^{t-1}$ to help the model remember the old tasks. 
\begin{equation}
    \mathcal{L}_{pre} = \text{KL}(\hat{\vy}_\mathcal{C}, \hat{\vy}_\mathcal{S}) + \text{MSE}(\hat{\vf}_\mathcal{C}, \hat{\vf}_\mathcal{S})\;,
\end{equation}
where the old synthetic data/label pair $(\hat{\vx}, \hat{\vy}) \in \mathcal{M}^{t-1}$, $\hat{\vy}_\mathcal{C}, \hat{\vf}_\mathcal{C}$ are from $\mathcal{C}^t_k(\hat{\vx})$ and $\hat{\vy}_\mathcal{S}, \hat{\vf}_\mathcal{S}$ are from $\mathcal{S}^{t-1}(\hat{\vx})$. 
The $\text{KL}$ (Kullback-Leibler divergence) term is utilized to distill the logits of the previous model into the current model. Furthermore, we propose using the $\text{MSE}$ term to ensure that the feature embedding of the current model remains consistent with the previous one. This consistency helps organize the high-confidence regions of $\mathcal{C}^t_k$.

Combining the above losses, we obtain the loss of the client $\mathcal{C}^t_k$ as follows:
\begin{equation}
    \mathcal{L}_\mathcal{C} = \alpha_{cur}^t\mathcal{L}_{cur} + \alpha_{pre}^t\mathcal{L}_{pre}\;.
    \label{eq:lcl}
\end{equation}
Inspired by \cite{rdfcil}, to address the difficulty of preserving previous knowledge grows as the ratio of previous classes to new classes
gets larger, the scale factor $\alpha_{cur}^t$ and $\alpha_{pre}^t$ are adaptively set as follows:
\begin{align}
    \alpha_{cur}^t = \frac{1+1/\kappa}{\delta}\alpha_{cur}; \alpha_{pre}^t = \kappa\delta\alpha_{pre}\;,
\end{align}
\text{where} $\kappa = \log_2(\frac{|\mathcal{Y}^{t}|}{2} + 1)$, $\delta = \sqrt{\frac{|\mathcal{Y}^{1:t-1}|}{|\mathcal{Y}^{t}|}}$, $|\mathcal{Y}^{t}|$ is the number of classes in task $t$, $\alpha_{cur}$ and $\alpha_{pre}$ are the base factors. 
\subsection{Global-Side: LT-Centered Data Generation}
\label{sec:s1}
Data-free generation, a forefront tool in FCIL, models global data distribution without compromising client privacy. Techniques like \cite{target, fcl, flcm} utilize a pretrained server model $S^{t-1}$ and random noise $\vz \sim \mathcal{N}(0, I)$ via a generator $\mathcal{G}$ to craft synthetic images. However, a drawback is the use of random noise lacking meaningful information, resulting in low-quality samples and limitations in addressing the forgetting problem. Our approach, inspired by \cite{nayer}, leverages meaningful LTE as input to capture valuable interclass information, generating high-quality images swiftly. We introduce a noisy layer at the layer level, incorporating random noise to prevent overreliance on unchanging label information. Random reinitialization of the noisy layer with each iteration enhances diversity in synthesized images, effectively mitigating the risk of overemphasizing label information.

First, we randomly sample the pseudo label $\hat{\vy}$ from a categorical distribution, and then we consider the LTEs of $\hat{\vy}$ query from $\mathcal{P}$ (i.e. $\ve_{\hat{\vy}} \sim \mathcal{P}$) as the input for the noisy layer $\mathcal{Z}$. Subsequently, the output of $\mathcal{Z}(\ve_{\hat{\vy}})$ is fed into the generator $\mathcal{G}$ to produce a batch of synthetic images $\hat{\vx}$.
\begin{align}
    \hat{\vx} &= \mathcal{G}(\mathcal{Z}(\ve_{\hat{\vy}}))\;,
    \label{eq:xg}
\end{align}
where $\mathcal{Z}$ is designed as the combination of a \code{BatchNorm} layer and a single \code{Linear} layer, follows to \cite{nayer}.

\noindent
\textbf{Learnable Data Stats}. In DFKT, reusing the training data stats as the normalization values for synthetic data is a common technique to generate high-quality images \cite{fastdfkd,target,nayer}. In these approaches, synthetic data $\hat{\vx}$ is normalized using the mean and standard deviation calculated from the entire training dataset. However, in a real-world federated setting, computing these stats can be challenging since the data are decentralized, and providing this information may raise privacy concerns. To address this issue, we propose using learnable data stats (LDS) with learnable mean $\mu$ and standard deviation $\sigma$, which are then trained together with $\mathcal{Z}$ and $\mathcal{G}$. From that, the synthetic data is generated as follows:
\begin{equation}
    \hat{\vx} = (\mathcal{G}(\mathcal{Z}(\ve_{\hat{\vy}})) - \mu)/\sigma\;.
\end{equation}
We also conducted experiments to demonstrate the benefits of using LDS, as discussed in Section \ref{sec:ab}. Without requiring the client's data stats, LDS still achieves comparable performance to the use of training data stats and significantly outperforms methods using a random stats or those without normalization technique.

Further, to effectively perform knowledge transfer, these synthetic data need to provide three essential properties including: \textit{Similarity, Stability} and \textit{Diversity}. 

\noindent
\textbf{Similarity.} The synthetic data $\hat{\vx}$ needs to be similar to the real training data. However, due to lack of access to the client data, we achieve this by minimizing the logits of $S^{t-1}$ and the pseudo label $\hat{\vy}$ through the following cross-entropy (CE) loss.
\begin{equation}
    \mathcal{L}_\mathcal{G}^{oh} = \text{CE}(\hat{\vy}_{\mathcal{S}^{t-1}}, \hat{\vy})\;, 
\end{equation}
where we denote $\hat{\vy}_{\mathcal{S}^{t-1}}$ as the prediction of $\mathcal{S}^{t-1}$ on $\hat{\vx}$.

\noindent
\textbf{Stability.} To enhance the generator's stability, we use the common batch normalization regularization \cite{adi,fastdfkd} to align the mean and variance of features at the batch normalization layer with their running counterparts.
\begin{equation}
    \mathcal{L}_\mathcal{G}^{bn} = \sum\limits_{l}\big(\|\mu_l(\hat{\vx}) -\mu_l\| + \|\sigma^2_l(\hat{\vx}) -\sigma^2_l\|\big)\;,
\end{equation}
where $\mu_l(\hat{\vx})$ and $\sigma^2_l(\hat{\vx})$ are the mean and variance of the $l$-th \code{BatchNorm} layer of $\mathcal{G}$, and $\mu_l$ and $\sigma^2_l$ are the mean and variance of the $l$-th \code{BatchNorm} layer of $\mathcal{S}^{t-1}$.

\noindent 
\textbf{Diversity.} To avoid the generation of similar images, LANDER incorporates an additional discriminator (student) network $\mathcal{Q}$. Specifically, for the synthetic images $\hat{\vx}$, $\mathcal{Q}$ is trained to minimize the difference between its predictions and those of the server (teacher) model.
\begin{equation}
    \mathcal{L}_\mathcal{Q} = \text{KL}(\hat{\vy}_\mathcal{Q}, \hat{\vy}_{\mathcal{S}^{t-1}}) + \text{MSE}(\hat{\vf}_\mathcal{Q}, \hat{\vf}_{\mathcal{S}^{t-1}})\;,
    \label{eq:ld}
\end{equation}
where $\hat{\vy}_\mathcal{Q}, \hat{\vf}_\mathcal{Q}$ are from $\mathcal{Q}(\hat{\vx})$.

On the other hand, by minimizing the negative KL Loss in Eq. \ref{eq:gadv}, the generator is designed to produce images that the student network has not learned before. 
\begin{align}
    \mathcal{L}_\mathcal{G}^{adv} &= -\omega \text{KL}(\hat{\vy}_\mathcal{Q}, \hat{\vy}_{\mathcal{S}^{t-1}})\;,\\
    \omega &= \mathbbm{1}(\code{argmax}(\hat{\vy}_{\mathcal{S}^{t-1}}) \neq \code{argmax}(\hat{\vy}_\mathcal{Q}))\;,
    \label{eq:gadv}
\end{align}
where $\mathbbm{1}(P)$ yields 1 if $P$ is true and 0 if $P$ is false. This optimization process enables the generation of a diverse set of images that cover the entire high-confidence space of the previous server model.

Furthermore, to utilize the capabilities of the anchored latent space as discussed in Section \ref{sec:c1}, we propose to use the Bounding Loss (Eq. \ref{eq:b}) to ensure that the synthetic images remain within the LTEs area. This ensures that these data contain more meaningful and helpful information for transferring knowledge from the previous server model.
\begin{equation}
    \mathcal{L}_\mathcal{G}^{ltc} = \text{B}(\hat{\vf}_{\mathcal{S}^{t-1}}, \ve_{\hat{\vy}})\;.
\end{equation}
In summary, the final objective for the generator $\mathcal{G}$, noisy layer $\mathcal{Z}$, learnable mean $\mu$ and standard deviation $\sigma$ is provided as followed:
\begin{equation}
    \mathcal{L}_{\mathcal{G},\mathcal{Z},\mu,\sigma} = \mathcal{L}_\mathcal{G}^{adv} +\lambda_{bn}\mathcal{L}_\mathcal{G}^{bn}  + \lambda_{oh}\mathcal{L}_\mathcal{G}^{oh} + \lambda_{ltc}\mathcal{L}_\mathcal{G}^{ltc}\;.
    \label{eq:lzg}
\end{equation}
Specifically, we set $\lambda_{bn}=1.0$, $\lambda_{oh}=0.5$, and $\lambda_{ltc}=5$ in all experiments.

\section{Experiments}

\begin{table*}[t]
\caption{The average accuracy (\%) and forgetting values over 3 trials for all learned tasks on CIFAR-100 are presented for different task numbers (5, 10) under both IID and non-IID settings. NIID($\beta$) indicate the Dirichlet parameter is set to $\beta$, 'Acc' denotes average accuracy, and '$\mathcal{F}$' signifies the forgetting measure \cite{forget1,forget2}. The results of Finetune, FedEWC, FedWeIT, FedLwF and TARGET are from \cite{target} and the best results are highlighted in bold. It's important to note that we exclude the results of MFCL \cite{mfcl} and Fed-CIL \cite{fedcil} due to their reporting in a significantly different setting, and their source codes are unavailable. }
\begin{adjustbox}{width=\linewidth}
\begin{tabular}{@{}lcccccccccccccccc@{}}
\toprule
 &
  \multicolumn{8}{c}{Acc(↑)} &
  \multicolumn{8}{c}{$\mathcal{F}$(↓)} \\ \midrule
Data   partition &
  \multicolumn{2}{c}{IID} &
  \multicolumn{2}{c}{NIID (1)} &
  \multicolumn{2}{c}{NIID (0.5)} &
  \multicolumn{2}{c}{NIID (0.1)} &
  \multicolumn{2}{c}{IID} &
  \multicolumn{2}{c}{NIID (1)} &
  \multicolumn{2}{c}{NIID (0.5)} &
  \multicolumn{2}{c}{NIID (0.1)} \\
Tasks &
  \multicolumn{1}{c}{T=5} &
  \multicolumn{1}{c}{T=10} &
  \multicolumn{1}{c}{T=5} &
  \multicolumn{1}{c}{T=10} &
  \multicolumn{1}{c}{T=5} &
  \multicolumn{1}{c}{T=10} &
  \multicolumn{1}{c}{T=5} &
  \multicolumn{1}{c}{T=10} &
  \multicolumn{1}{c}{T=5} &
  \multicolumn{1}{c}{T=10} &
  \multicolumn{1}{c}{T=5} &
  \multicolumn{1}{c}{T=10} &
  \multicolumn{1}{c}{T=5} &
  \multicolumn{1}{c}{T=10} &
  \multicolumn{1}{c}{T=5} &
  \multicolumn{1}{c}{T=10} \\
Finetune &
  16.12 &
  7.83 &
  16.33 &
  8.45 &
  15.49 &
  7.64 &
  - &
  - &
  78.12 &
  75.89 &
  77.59 &
  74.89 &
  74.95 &
  71.52 &
  - &
  - \\
FedEWC \cite{forget2} &
  16.51 &
  8.01 &
  16.06 &
  8.84 &
  16.86 &
  8.04 &
  - &
  - &
  71.12 &
  65.06 &
  68.02 &
  62.14 &
  62.40 &
  65.23 &
  - &
  - \\
FedWeIT \cite{fclwt} &
  28.45 &
  20.39 &
  28.56 &
  19.68 &
  24.57 &
  15.45 &
  - &
  - &
  52.12 &
  43.18 &
  49.84 &
  45.82 &
  45.96 &
  48.54 &
  - &
  - \\
FedLwF \cite{lwf} &
  30.61 &
  23.27 &
  30.94 &
  21.16 &
  27.59 &
  17.98 &
  - &
  - &
  45.32 &
  37.71 &
  42.71 &
  41.03 &
  41.25 &
  45.23 &
  - &
  - \\
TARGET \cite{target} &
  36.31 &
  24.76 &
  34.89 &
  22.85 &
  33.33 &
  20.71 &
  28.32 &
  19.25 &
  32.23 &
  35.45 &
  34.48 &
  38.25 &
  39.23 &
  42.23 &
  38.23 &
  45.23 \\
LANDER (Ours) &
  \textbf{52.60} &
  \textbf{40.21} &
  \textbf{51.78} &
  \textbf{37.21} &
  \textbf{48.23} &
  \textbf{33.35} &
  \textbf{43.42} &
  \textbf{29.29} &
  \textbf{18.03} &
  \textbf{25.56} &
  \textbf{18.92} &
  \textbf{28.92} &
  \textbf{30.61} &
  \textbf{32.86} &
  \textbf{15.20} &
  \textbf{28.69} \\ \bottomrule
\end{tabular}
\end{adjustbox}
\label{tab:cifar100}
\end{table*}

\subsection{Experimental Setting}
We conduct experiments on CIFAR-100 \cite{c10}, Tiny-ImageNet \cite{tin}, and ImageNet \cite{in} to evaluate our proposed approach. Following to \cite{cl_gr, target, mfcl}, we partition the dataset classes into tasks, mimicking class continual learning with 5 and 10 tasks. ResNet18 \cite{resnet} serves as the backbone for all experiments.

The evaluation employs traditional CL metrics, including average accuracy and a forgetting score \cite{forget1, forget2}. Following to \cite{target}, our approach is compared against four baseline types in FCIL: 1) Finetune, which learns each task on each client sequentially; 2) FedWeIT \cite{fclwt}, a widely used regularization-based method; 3) FedEWC \cite{forget2} and FedLwF \cite{lwf}, the application of common data-free CL techniques in the federated scenario; 4) TARGET \cite{target}, a current SOTA FCIL method. For details on task configuration, hyper-parameters, additional results, and visualizations, please refer to the \textbf{Appendix}.

\subsection{Main Results}
\begin{figure}[t]
\begin{center}
\includegraphics[width=0.65\linewidth]{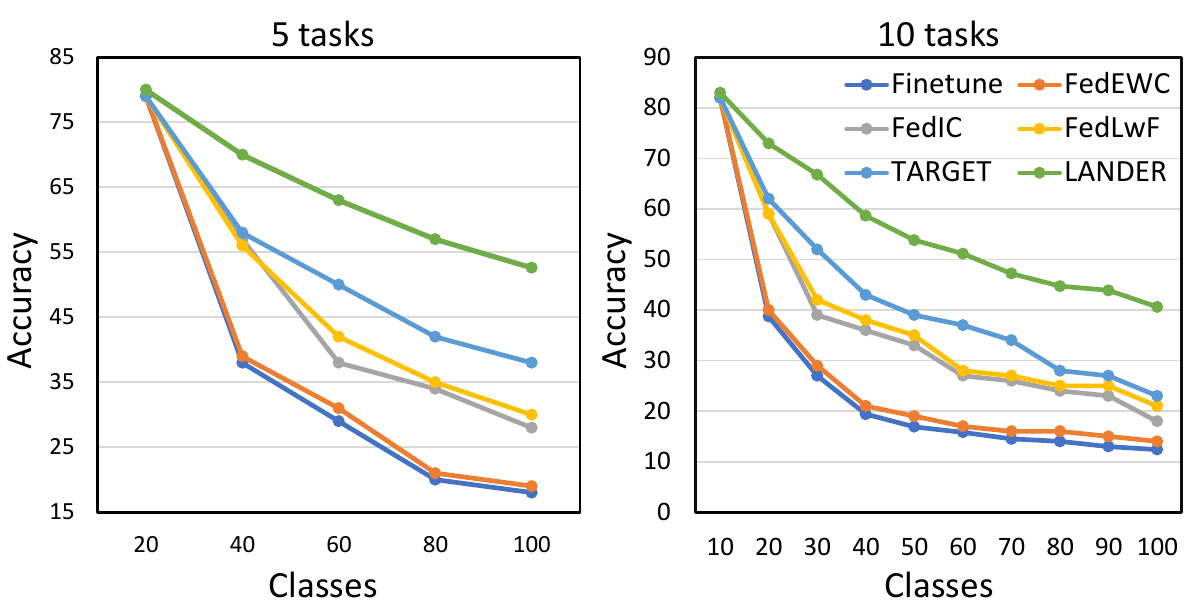}
\end{center}
\caption{Average accuracy on incremental tasks.}
\label{fig:clacc}
\end{figure}

\noindent
\textbf{Experiments on CIFAR100.} Following \cite{target}, we conduct experiments on 5 and 10 tasks in both IID and non-IID scenarios. Table \ref{tab:cifar100} shows the final average accuracy of the server model and the forgetting measure for each experiment. Notably: 1) LANDER outperforms prior methods by two digits in both accuracy and forgetting score in all settings; 2) Increasing tasks and lowering Dirichlet parameters substantially reduce accuracy for compared methods; 3) Even in highly skewed settings, LANDER consistently outperforms others, highlighting its superior performance. Figure \ref{fig:clacc} illustrates our model's superior performance in all incremental tasks.  The graph depicts average accuracy on current and previous tasks, emphasizing the model's effectiveness in enabling local clients to learn new classes in a streaming manner while mitigating forgetting.

\noindent
\textbf{Experiments on Large-scale Datasets.} To evaluate LANDER's effectiveness, we conducted additional assessments on the more challenging Tiny-ImageNet and ImageNet datasets. For the Tiny-ImageNet dataset (Table \ref{tab:tiny}), we present final average accuracy and forgetting measures for all tasks in both IID and non-IID settings with 5 tasks. The results show our method consistently achieves an approximately 3\% higher average accuracy than FedLwF and TARGET. Additionally, our method exhibits significantly lower forgetting measures than FedLwF in both settings, highlighting its effectiveness in mitigating catastrophic forgetting in the presence of extreme data distributions.

For ImageNet (Table \ref{tab:in}), as most FCIL methods do not report results on this dataset, our comparison primarily involves the current SOTA FCIL method, TARGET \cite{target}. For a fair comparison, we re-conducted TARGET's experiments to align with our settings. The results clearly demonstrate LANDER outperforms other methods in accuracy, showcasing its efficacy on a large-scale dataset.

\begin{table}[]
\caption{The Average Accuracy (\%) and Forgetting for all learned tasks on Tiny-ImageNet for 5 tasks.}
\begin{adjustbox}{width=\linewidth}
\begin{tabular}{@{}clccccc@{}}
\toprule
\multicolumn{1}{l}{} &
  Data   partition &
  \multicolumn{1}{c}{IID} &
  \multicolumn{1}{c}{NIID(1)} &
  \multicolumn{1}{c}{NIID(0.5)} &
  \multicolumn{1}{c}{NIID(0.1)} &
  \multicolumn{1}{c}{NIID(0.05)} \\ \midrule
\multirow{3}{*}{Acc(↑)} & FedLwF \cite{lwf}        & 24.32          & 22.56          & 21.76          & 18.78          & 18.59          \\
                        & TARGET \cite{target}        & 26.25          & 24.12          & 23.95          & 21.15          & 20.95          \\
                        & LANDER (Ours) & \textbf{30.29} & \textbf{28.21} & \textbf{27.98} & \textbf{25.27} & \textbf{25.02} \\ \midrule
\multirow{3}{*}{$\mathcal{F}$(↓)}   & FedLwF \cite{lwf}       & 34.57          & 33.94          & 37.23          & 31.43          & 31.19          \\
                        & TARGET \cite{target}        & 23.43          & 25.12          & 24.58          & 20.54          & 20.83          \\
                        & LANDER (Ours) & \textbf{21.65} & \textbf{23.09} & \textbf{23.03} & \textbf{17.93} & \textbf{18.14} \\ \bottomrule
\end{tabular}
\end{adjustbox}
\label{tab:tiny}
\end{table}

\begin{table}[]
\caption{The Average Accuracy (\%) on ImageNet for 5 tasks.}
\begin{adjustbox}{width=\linewidth}
\begin{tabular}{@{}clccccc@{}}
\toprule
\multicolumn{1}{l}{}     & Method & Task 1         & Task 2         & Task 3         & Task 4         & Task 5         \\ \midrule
\multirow{2}{*}{IID}     & TARGET \cite{target} & 77.16          & 55.32      	& 45.67	         & 36.19	      & 31.83         \\
                         & LANDER (Ours) & \textbf{77.32} & \textbf{65.42} & \textbf{56.34} & \textbf{48.82} & \textbf{43.24} \\ \midrule
\multirow{2}{*}{NIID(1)} & TARGET \cite{target} & 76.78          & 54.74	        & 42.67	         & 31.19	      & 29.83          \\
                         & LANDER (Ours) & \textbf{76.91} & \textbf{63.35} & \textbf{54.25} & \textbf{45.35} & \textbf{41.75} \\ \bottomrule
\end{tabular}
\end{adjustbox}
\label{tab:in}
\end{table}

\subsection{Ablations Studies}
\label{sec:ab}
\begin{figure}[t]
\begin{center}
\includegraphics[width=1\linewidth]{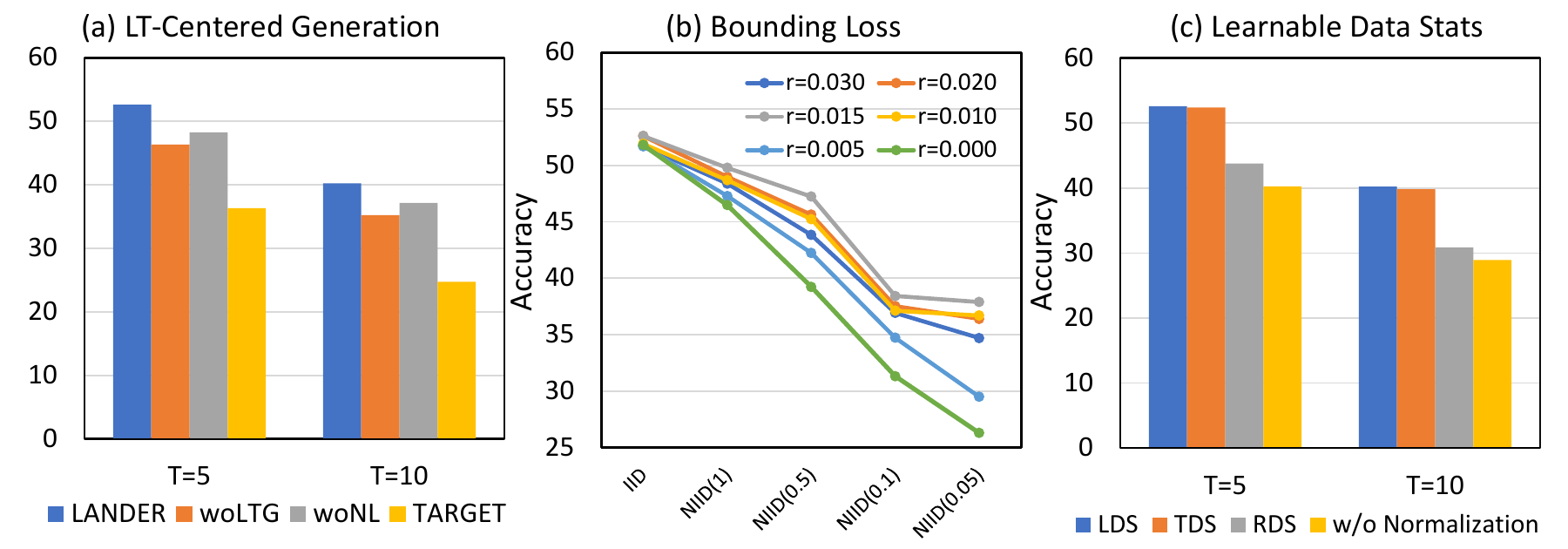}
\end{center}
\caption{(a) Accuracies in 5 and 10 tasks of our method with and without LT-centered generation. (b) Incremental Accuracy on CIFAR-100 for different values of $r$ in Bounding Loss. (c) Accuracies for different kinds of data normalization values.}
\label{fig:ab_bl_ln}
\end{figure}

\noindent
\textbf{Effectiveness of LT-Centered Generation}. In Figure \ref{fig:ab_bl_ln}a, we analyze the impact of removing LTE constraints in data-free generation (woLTG) and the combination of a noisy layer and LTE as input (woNL) on our methods. The results indicate that these two components play important roles in our LANDER.

\noindent
\textbf{Effectiveness of Bounding Loss.} In Figure \ref{fig:ab_bl_ln}b, we assess the impact of different radius $r$ in the Bounding Loss within a heterogeneous setting. The results demonstrate that: 1) our method achieves the best results with a radius of $r$=0.015, which is approximately half of the minimum class-wise L2 distance in CIFAR100; 2) with a higher Dirichlet parameter, the method without using Bounding Loss ($r$=0) exhibits significantly lower accuracy compared to the one using $r$=0.015, highlighting the benefits of our Bounding Loss in addressing the imbalance problem in federated learning.

\noindent
\textbf{Effectiveness of Learnable Data Stats.} In Figure \ref{fig:ab_bl_ln}c, we compare the normalization using our LDS with that using training data stats (TDS) \cite{target, fedcil}, random data stats (RDS), and without using normalization. The results show: 1) Without normalization or with random data stats, our work's performance decreases significantly; 2) Our LDS performs comparably or slightly better than the TDS for data normalization, demonstrating its benefits in our work.

\subsection{Further Analysis}
\noindent
\textbf{Comparison with Different Prompting Engineering Templates.} We assess the influence of various prompting engineering techniques for generating label text. We introduce three approaches to prompt label text: P1: \code{"a class of a \{class\_name\}"}, P2: \code{"a photo of a \{class\_name\}"}, P3: \code{"a photo of a \{class\_index\}"}. Table \ref{tab:as_prompt} highlights that P2 outperform the best in the comparison. Furthermore, even when using only the label index, P3 maintains a performance advantage over the best baseline. This indicates the application of using label indices in datasets with less meaningful labels, highlighting the practical effectiveness of LANDER.

\begin{table}[H]
\centering
\caption{Accuracies of different prompt engineering methods.}
\begin{adjustbox}{width=1\linewidth}
\begin{tabular}{@{}lcccccccc@{}}
\toprule
             & \multicolumn{4}{c}{IID} & \multicolumn{4}{c}{NIID(0.5)} \\ \midrule
Text Encoder & SOTA & P1    & P2    & P3  &  SOTA& P1   & P2   & P3    \\
Accuracy           & 36.31 & 52.45 & \textbf{52.60}     & 50.12     & 33.33 & 48.09 & \textbf{48.23}  & 45.39      \\ \bottomrule
\end{tabular}
\end{adjustbox}
\label{tab:as_prompt}
\end{table}

\noindent
\textbf{Comparison with Different Text Encoder.} We evaluate our LANDER across three common text encoders: Doc2Vec \cite{doc2vec}, SBERT \cite{sbert}, and CLIP \cite{clip}. Table \ref{tab:as_multe} reveals that LANDER performs well across diverse language models, leveraging their capacity to capture label-text relations. When coupled with the label-index prompt engineering method discussed earlier, our approach adapts effectively to different domains, even without meaningful label-text. Furthermore, utilizing foundational models like CLIP enhances our model's performance marginally, this indicate the benefit of multimodal models to our works. Finally, we select CLIP as the text encoder for this paper.

\begin{table}[H]
\centering
\caption{Accuracies of our LANDER with different LM.}
\begin{adjustbox}{width=1\linewidth}
\begin{tabular}{@{}lcccccccc@{}}
\toprule
             & \multicolumn{4}{c}{IID} & \multicolumn{4}{c}{NIID(0.5)} \\ \midrule
Text Encoder & SOTA & Doc2Vec    & SBERT    & CLIP  & SOTA & Doc2Vec   & SBERT   & CLIP    \\
Accuracy           & 36.31 & 52.45     & 52.53   & \textbf{52.60}  & 33.33 & 48.21    & 48.18   & \textbf{48.23}   \\ \bottomrule
\end{tabular}
\end{adjustbox}
\label{tab:as_multe}
\end{table}

\begin{figure}[t]
\begin{center}
\includegraphics[width=0.95\linewidth]{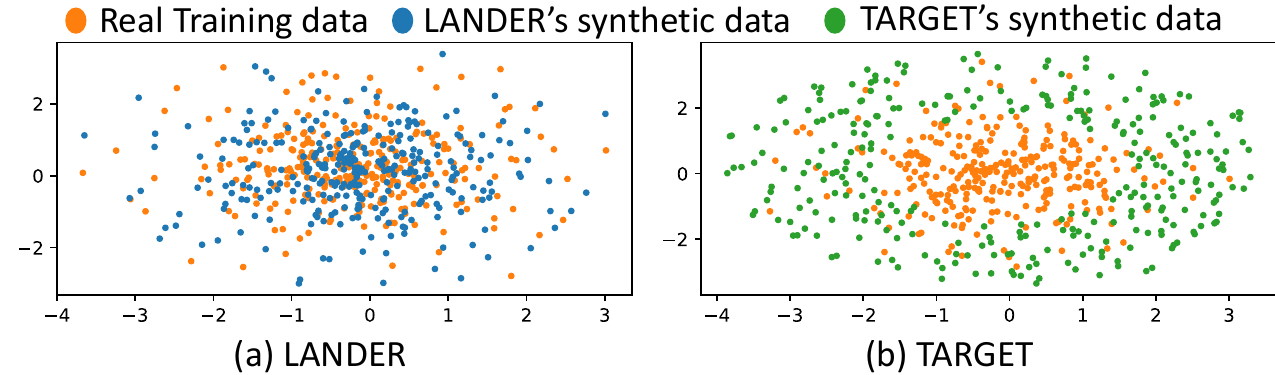}
\end{center}
\caption{Visualizing t-SNE on synthetic and real data in a randomly chosen class in CIFAR-100. Our LANDER generates data in latent space that closely resembles real data.}
\label{fig:tsne}
\end{figure}

\noindent
\textbf{Visualization}. The t-SNE visualization in Figure \ref{fig:tsne} illustrates synthetic data generated by our LANDER and TARGET. It is evident that, using the anchor LTE, LANDER generates samples with embeddings more similar to real data compared to TARGET \cite{target}. This observation underscores the reason behind the improvement in our approach.

\begin{figure}[t]
\begin{center}
\includegraphics[width=1\linewidth]{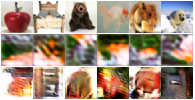}
\end{center}
\caption{Real data (top) vs synthetic data generated by TARGET (middle) and our LANDER (bottom) in CIFAR-100. Each column shows samples from the same class.}
\label{fig:visual}
\end{figure}

\noindent
\textbf{Visualization on Synthetic Data}. Figure \ref{fig:visual} compares real and synthetic data from LANDER and TARGET. Our synthetic samples intentionally differ from specific training examples, preserving privacy. Unlike TARGET's meaningless images, our samples capture essential class knowledge and effectively represent the entire class. Consequently, the inclusion of synthetic samples significantly mitigates the issue of catastrophic forgetting.

\section{Conclusion}
In this paper, we propose LANDER to mitigate the forgetting issue in federated learning.  Specifically, we treat LTEs as anchor points of data feature embeddings during model training, enriching surrounding feature embeddings. In the DFKT phase, LANDER leverages thse LTE anchors to synthesize more meaningful samples, effectively addressing forgetting.  Extensive experimental results demonstrate that our method achieves SOTA performance in FCIL. 

\section*{Acknowledgements}
This work was supported by ARC DP23 grant DP230101176 and by the Air Force Office of Scientific Research under award number FA2386-23-1-4044.

{\small
\bibliographystyle{ieee_fullname}
\bibliography{egbib}
}

\appendix
\section{Experimental Detail}

\subsection{Evaluation Metric}

In this works, we used two commonly used evaluation metric in FCIL including the averaging accuracy and averaging forgetting score.
\begin{itemize}[noitemsep, nolistsep]
    \item Last Incremental Averaging Accuracy (Acc): It is the accuracy of server model after training all tasks.
    \item Averaging Forgetting ($\mathcal{F}$): $\mathcal{F}^t$ of task $t$ is defined as the difference between the highest accuracy of the model on task $t$ and its performance at the end of the training. Therefore, we can evaluate the average forgetting $\mathcal{F}$ by averaging all the $\mathcal{F}^t$ for task 1 to $T$ - 1 at the end of task $T$.
\end{itemize}

\subsection{Client Training Details}

For a fair comparison, in accordance with \cite{target}, we employed ResNet18 \cite{resnet} as the backbone for all experiments. Each client was trained with a batch size of 128 for 100 communication rounds, with 2 local training epochs per communication round. For CIFAR-100, we used the SGD optimizer with a learning rate of 0.04, a momentum of 0.9, and a weight decay of 5e-4. In the case of Tiny-ImageNet and ImageNet, a learning rate of 0.1, weight decay of 2e-4, and a multi-step scheduler were applied, reducing the learning rate by 10 at the 50th and 75th communication rounds.

\subsection{Generator Network Training Details}

The generator architecture for CIFAR-100 and Tiny-ImageNet is detailed in Table \ref{tab:g_arch}, while the generator architecture for ImageNet is presented in Table \ref{tab:g_arch_in}. We utilize the Adam optimizer with a learning rate of 2e-3 to optimize the generator. For CIFAR-100 and Tiny-ImageNet, we configure the synthetic batch size to be 256, whereas for ImageNet, the synthetic batch size is set to 128 to ensure that GPU memory remains below 24GB.

\begin{table}[h]
\centering
\caption{Generator Network ($\mathcal{G}$) Architecture for CIFAR-100 and Tiny-ImageNet.}
\begin{adjustbox}{width=\linewidth}
\begin{tabular}{@{}ll@{}}
\toprule
Output          & \textbf{Size   Layers}                              \\ \midrule
256            & Input                                 \\
$128 \times h/4 \times w/4$ & Linear, BatchNorm1D, Reshape                        \\
$128 \times h/4 \times w/4$ & SpectralNorm (Conv (3 × 3)), BatchNorm2D, LeakyReLU \\
$128 \times h/2 \times w/2$ & UpSample (2×)                                       \\
$64 \times h/2 \times w/2$  & SpectralNorm (Conv (3 × 3)), BatchNorm2D, LeakyReLU \\
$64 \times h \times w$     & UpSample (2×)                                       \\
$3 \times h \times w$      & SpectralNorm (Conv (3 × 3)), Sigmoid, BatchNorm2D      \\ \bottomrule
\end{tabular}
\end{adjustbox}
\label{tab:g_arch}
\end{table}
\begin{table}[h]
\centering
\caption{Generator Network ($\mathcal{G}$) Architecture for ImageNet.}
\begin{adjustbox}{width=\linewidth}
\begin{tabular}{@{}ll@{}}
\toprule
Output          & \textbf{Size   Layers}                              \\ \midrule
256            & Input                                 \\
$128 \times h/16 \times w/16$ & Linear, BatchNorm1D, Reshape                        \\
$128 \times h/16 \times w/16$ & SpectralNorm (Conv (3 × 3)), BatchNorm2D, LeakyReLU \\
$128 \times h/8 \times w/8$ & UpSample (2×) \\                    
$128 \times h/8 \times w/8$ & SpectralNorm (Conv (3 × 3)), BatchNorm2D, LeakyReLU \\
$128 \times h/4 \times w/4$ & UpSample (2×) \\                      
$64 \times h/4 \times w/4$ & SpectralNorm (Conv (3 × 3)), BatchNorm2D, LeakyReLU \\
$64 \times h/2 \times w/2$ & UpSample (2×) \\
$64 \times h/2 \times w/2$  & SpectralNorm (Conv (3 × 3)), BatchNorm2D, LeakyReLU \\
$64 \times h \times w$     & UpSample (2×)                                       \\
$3 \times h \times w$      & SpectralNorm (Conv (3 × 3)), Sigmoid, BatchNorm2D      \\ \bottomrule
\end{tabular}
\end{adjustbox}
\label{tab:g_arch_in}
\end{table}

\subsection{Hyperparameters Tuning}
In this section, we present the hyperparameter tuning approaches and results used in our experiments. The hyperparameters for CIFAR-100, Tiny-ImageNet, and ImageNet datasets are summarized in Table \ref{tab:hyperpara}.

\begin{table}[h]
\caption{The hyperparameters for LANDER applied to CIFAR-100, Tiny-ImageNet, and ImageNet are detailed below. Specifically, $\alpha_{cur}$ and $\alpha_{pre}$ are the base factors for training the client model, while $\lambda_{cls}$, $\lambda_{bn}$, and $\lambda_{adv}$ are the hyperparameters associated with generative model training. The variables $g$ represent the training steps to optimize the generator, and $I$ is the number of rounds for generating synthetic images.}
\begin{adjustbox}{width=1\linewidth}
\begin{tabular}{|l|c|c|c|c|c|l|l|}
\hline
         & \multicolumn{1}{l|}{$\alpha_{cur}$} & \multicolumn{1}{l|}{$\alpha_{pre}$} & \multicolumn{1}{l|}{$\lambda_{bn}$} & \multicolumn{1}{l|}{$\lambda_{oh}$} & \multicolumn{1}{l|}{$\lambda_{ltc}$} &  $g$  & $I$  \\ \hline
CIFAR-100 & \multirow{3}{*}{0.2}     & \multirow{3}{*}{0.4} & \multirow{3}{*}{1}      & \multirow{3}{*}{0.5}    & \multirow{3}{*}{5}            & 40 & 40 \\ \cline{1-1} \cline{7-8} 
Tiny-ImageNet &  &  &  &  &  & 100 & 200 \\ \cline{1-1} \cline{7-8} 
ImageNet     &  &  &  &  &  & 200 & 400 \\ \hline
\end{tabular}
\end{adjustbox}
\label{tab:hyperpara}
\end{table}

\noindent
\textbf{Scale Factor Tuning}. Scale factor hyperparameters play a crucial role in algorithm performance. In our experiments, we demonstrate the sensitivity of the final performance to each hyperparameter. Due to the computational expense of hyperparameter tuning in this setting, we systematically vary one parameter at a time while keeping others fixed. The final accuracy for CIFAR-100 datasets with five tasks and a Dirichlet parameter set at 0.5 is reported in Table \ref{tab:gsf}.

\begin{table}[H]
\centering
\caption{We examine the impact of various hyperparameters on the final accuracy for CIFAR-100. Here, $\alpha_{cur}$ and $\alpha_{pre}$ denote the scale factors for client model training, while $\lambda_{bn}$ and $\lambda_{oh}$ serve as the scale factors for generator training. Additionally, $\lambda_{ltc}$ is utilized for both client and generator training.}
\begin{adjustbox}{width=\linewidth}
\begin{tabular}{@{}llllllllll@{}}
\toprule
 $\alpha_{cur}$ & Acc   &  $\alpha_{pre}$ & Acc & $\lambda_{bn}$   & Acc & $\lambda_{oh}$  & Acc & $\lambda_{ltc}$ & Acc \\ \midrule
0.1 & 47.12 & 0.1 & 46.37    & 0.01 & 46.25    & 0.1 & 47.79    & 0.5 & 47.93    \\
0.2 & 48.23 & 0.2 & 47.24    & 0.1  & 47.68    & 0.5 & 48.23    & 1   & 48.12    \\
0.4 & 46.92 & 0.4 & 48.23    & 1    & 48.23    & 1   & 48.12    & 5   & 48.23    \\
0.6 & 44.23 & 0.6 & 47.82    & 10   & 47.31    & 5   & 48.07    & 10  & 48.07    \\ \bottomrule
\end{tabular}
\end{adjustbox}
\label{tab:gsf}
\end{table}

\noindent
\textbf{Generation Training Steps $g$}. Due to the distinct challenges and image resolutions of CIFAR-100, Tiny-ImageNet, and ImageNet, we assess the impact of different generator training steps, denoted as $g$, for each dataset. Our findings reveal that the most crucial impact of $g$ lies in distilling knowledge from the server to the client. This is evaluated through the distilling accuracy of an additional student (discriminator) in generator training after the first task. Table \ref{tab:gts} illustrates that student accuracy perfectly aligns with the final accuracy across different $g$ values. Consequently, we propose using this metric to expedite parameter tuning.

Table \ref{tab:gts} shows that a low value of $g$ has negative effects on the performance of our work. Increasing the value of $g$ improves performance; however, an excessively high $g$ does not guarantee higher performance.

\begin{table}[H]
\centering
\caption{The impact of different generator training steps, denoted as $g$, on the final accuracy for three datasets. Here, "1st Acc" represents the distilling accuracy of the student model after the first task, while "Acc" signifies the final accuracy of the model.}
\begin{adjustbox}{width=\linewidth}
\begin{tabular}{@{}lll|lll|lll@{}}
\toprule
\multicolumn{3}{c}{CIFAR100}        & \multicolumn{3}{c}{Tiny-ImageNet}             & \multicolumn{3}{c}{ImageNet} \\ \midrule
g  & 1st Acc & Acc            & g            & 1st Acc & Acc            & g   & 1st Acc & Acc   \\
30 & 72.12           & 49.21          & 50           & 49.12           & 27.47          & 100 & 58.12           & 40.38 \\
40 & \textbf{74.61} & \textbf{51.78} & \textbf{100} & \textbf{52.82} & \textbf{28.21} & \textbf{200} & \textbf{61.86} & \textbf{41.75} \\
50 & 74.51           & 51.71 & 150 & 52.19           & 28.12 & 300 & 61.37           & 41.65 \\
60 & 74.27           & 51.74          &              &                 &                &     &                 &       \\ \bottomrule
\end{tabular}
\end{adjustbox}
\label{tab:gts}
\end{table}

\noindent
\textbf{Data Synthetic Rounds $I$}. Building on the findings from the last section, we also utilize the distilling accuracy of the student model at the first task as the evaluation metric for tuning the different data synthetic rounds, denoted as $I$, for three datasets. The results from Table \ref{tab:dsr} indicate that with a low volume of synthetic data, our method's accuracy fails to provide sufficient information for effective learning from previous tasks. Increasing the data volume effectively mitigates the forgetting phenomenon and enhances performance. However, a continuous increase in data volumes does not yield a significant improvement in the model's performance. It's crucial to highlight that the data volume alone does not guarantee the effectiveness of synthetic data in enhancing machine learning models.

\begin{table}[H]
\centering
\caption{The effect of different data synthetic rounds, denoted as $I$, on the final accuracy for three datasets. In this context, "1st Acc" represents the distilling accuracy of the student model after the first task, and "SynData" denotes the total number of synthetic data. Please note that we set the synthetic batch size at 256 for CIFAR-100 and Tiny-ImageNet, while the synthetic batch size is 128 for ImageNet to keep the GPU memory below 24GB. }
\begin{adjustbox}{width=\linewidth}
\begin{tabular}{lll|lll|lll}
\toprule
\multicolumn{3}{c}{CIFAR100}  & \multicolumn{3}{c}{Tiny-ImageNet} & \multicolumn{3}{c}{ImageNet}    \\ \midrule
I  & SynData & 1st Acc & I    & SynData  & 1st Acc  & I   & SynData & 1st Acc \\
20 & 5120    & 73.31           & 100  & 25600    & 48.12            & 200 & 25600   & 57.42           \\
\textbf{40} & \textbf{10240} & \textbf{74.61} & 150 & 38400 & 50.95 & 300          & 38400 & 61.16          \\
60 & 15360          & 74.47 & \textbf{200} & \textbf{51200}  & \textbf{52.82} & \textbf{400} & \textbf{51200} & \textbf{61.86} \\
80 & 20480   & 74.32           & 250  & 64000    & 51.43            & 500 & 64000   & 60.82           \\ \bottomrule
\end{tabular}
\end{adjustbox}
\label{tab:dsr}
\end{table}
\begin{figure*}[t]
    \centering
    \includegraphics[width=0.96\linewidth]{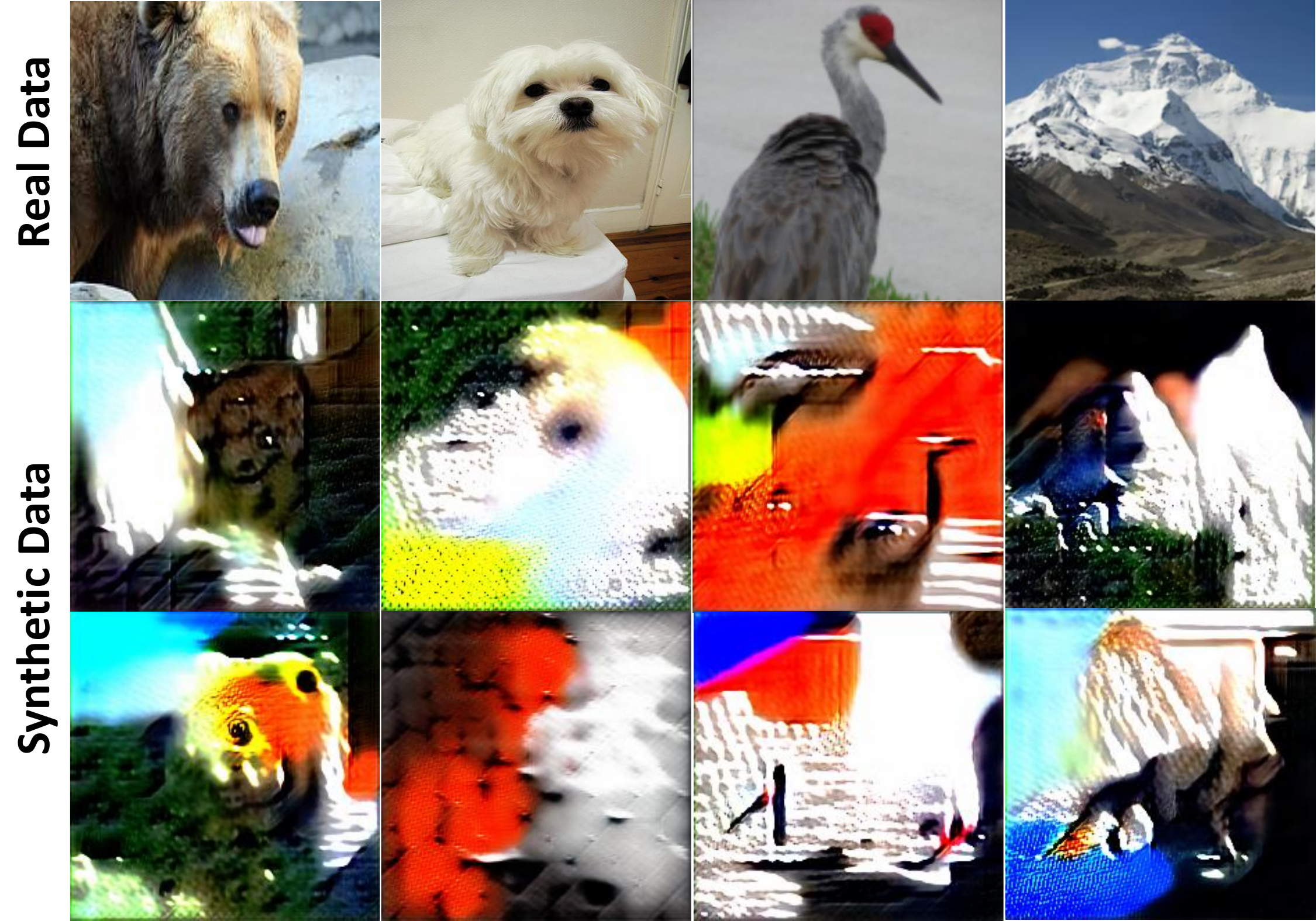}
    \caption{The top row displays real data, while the middle and bottom rows illustrate randomly generated synthetic data from our LANDER for ImageNet datasets.}
    \label{fig:imagenet-ex}
\end{figure*}

\section{Extended Results}
\noindent
\textbf{Discussing about Local Training Epoch.} We conducted experiments with different client training epochs in our methods and found that our approach is effective with epochs higher than 1. For optimal speed, we chose $l=2$, making our methods typically twice as fast as TARGET (4.25 to 9.87 hours), which requires local training epochs of 5. This demonstrates our method's superior performance and substantial reduction in training time compared to TARGET.

\begin{table}[H]
\caption{Accuracies and training time of our LANDER  over different local training epoches. This experiments is conducted using a single NVIDIA RTX 4090 in CIFAR-100 dataset with NIID(0.5).}
\begin{adjustbox}{width=\linewidth}
\begin{tabular}{@{}llcccccc@{}}
\toprule
       &              & l = 1 & l = 2 & l = 3 & l = 4 & l = 5 & l = 6 \\ \midrule
Method & Time (hours) & 2.35  & 4.25  & 6.13  & 8     & 9.87  & 11.75 \\
TARGET & Acc          & 27.94 & 29.19 & 30.27 & 32.18 & 33.33 & 33.26 \\
LANDER & Acc          & 43.24 & 48.23 & 48.23 & 48.23 & 48.23 & 48.23 \\ \bottomrule
\end{tabular}
\end{adjustbox}
\label{tab:nlte}
\end{table}
\noindent
\textbf{Comparison with Different Number of Clients}. We conducted additional experiments to evaluate our method with a higher number of clients. Table \ref{tab:ab_nc} demonstrates that even with an increased number of clients, our method maintains good performance. Notably, with 50 clients, our method significantly outperforms state-of-the-art methods (28.42\% compared to 12.72\%).

\begin{table}[H]
\caption{Comparison with Different Number of Clients}
\begin{adjustbox}{width=\linewidth}
\begin{tabular}{@{}lclllll@{}}
\toprule
       & \multicolumn{1}{l}{m=5} & m=10  & m=20  & m=30  & m=40  & m=50  \\ \midrule
TARGET & 36.31                   & 27.48 & 21.98 & 19.8  & 15.12 & 12.75 \\
LANDER & \textbf{52.60}          & 49.98 & 40.24 & 35.64 & 28.92 & 28.42 \\ \bottomrule
\end{tabular}
\end{adjustbox}
\label{tab:ab_nc}
\end{table}

\section{Privacy of LANDER}
Numerous attacks prevalent in federated learning, either in general or specifically in FedAvg \cite{fedcm}, include data poisoning, model poisoning, backdoor attacks, and gradient inversion attacks \cite{fls1, fls2, pri1, pri2}. In general, our approach does not introduce additional privacy concerns and maintains the same privacy issues as FedAvg. Therefore, our method is also compatible with any defense solutions used for FedAvg, such as secure aggregation \cite{secavg} or noise injection before aggregation \cite{sec2}.

In contrast to several existing FCIL approaches \cite{fedcil}, where clients need to share a locally trained generative model or perturbed private data, LANDER's generative model training relies on the weights of the global model, already shared with all clients in the FedAvg scenario. Anchoring the label-text embedding can enhance the data-free generation process while still preserving the privacy setting. Figure \ref{fig:imagenet-ex} illustrates several examples of synthetic images. From these examples, it is evident that the synthetic images exhibit significant differences compared to real data. However, they still encapsulate common knowledge and can be viewed as abstract visualizations of classes, thereby enhancing knowledge transfer.

Furthermore, we introduce Learnable Data Stats (LDS) to bolster the data privacy of the FCIL setting. Unlike prior methods \cite{fedcil, target, mfcl} that demand training data statistics, such as the mean and variance of all training data, for effective synthetic data generation, our approach with LDS achieves comparable performance without relying on these statistics. By eliminating the need for training data stats, our method enhances data privacy in the FCIL setting.

\section{Does LANDER Work in Meaningless Label Datasets?}

The paper proposes employing label-text embedding (LTE) while acknowledging its potential drawbacks in datasets lacking meaningful labels, such as those related to chemical compounds. Despite these challenges, our approach exclusively considers LTE as the optimal choice for anchoring each class. In instances of datasets with meaningless labels, our method can utilize the label index instead of the label description to generate the prompt, as demonstrated by \code{"a class of \{class\_index\}"}. Consequently, our method consistently outperforms the current state-of-the-art  approach when using only the label, as illustrated in Section 5.4 of the main paper.

Furthermore, our method can be compatible with any pretrained language model, thereby expanding the applications of our work in various real-world scenarios.

\section{Limitations and Future Works}
In our method, a primary limitation is the need to store and transfer a large amount of synthetic data from the server to the client, thereby increasing communication costs and the training load for the client. Consequently, reducing the volume of required synthetic data emerges as a potential direction for future work.

\end{document}

%% file: math_commands.tex
\usepackage{amsmath,amsfonts,bm}

\def\eqref#1{equation~\ref{#1}}

\def\1{\bm{1}}

\def\ve{{\bm{e}}}
\def\vf{{\bm{f}}}

\def\vx{{\bm{x}}}
\def\vy{{\bm{y}}}
\def\vz{{\bm{z}}}

\DeclareMathAlphabet{\mathsfit}{\encodingdefault}{\sfdefault}{m}{sl}
\SetMathAlphabet{\mathsfit}{bold}{\encodingdefault}{\sfdefault}{bx}{n}

